# AerialYield-B$^2$D: A Greenhouse Blueberry Dataset with Five-Stage Ripeness Masks and Fruit Counts

Iyyakutti Iyappan Ganapathi[1], Afeefa Azam[1], Muhammad Owais[2], Irfan Hussain[2], and Yusra Abdulrahman*[1,3]

[1]Department of Aerospace Engineering, Khalifa University, Abu Dhabi, United Arab Emirates
[2]Khalifa University Center for Autonomous Robotic Systems (KU-CARS), Khalifa University, Abu Dhabi, United Arab Emirates
[3]Advanced Research and Innovation Center (ARIC), Khalifa University, Abu Dhabi, United Arab Emirates

**Abstract**

Blueberry ripeness is judged by berry colour, cluster composition, and the distribution of maturity stages within a plant, however, public greenhouse image resources with dense ripeness-stage masks remain limited. We present AerialYield-B$^2$D, where B$^2$D denotes *BlueBerry Dataset*, a curated real-image resource containing 514 RGB images and 30,195 annotated blueberry instances across five ripeness stages: green immature, pale pink, pink-turns-purple, fully ripe and over-ripe. The release provides class-specific binary masks, overall berry masks, semantic label maps, image-level count tables, SHA-256 hashes, source metadata, recommended train/validation/test splits and technical validations. *AerialYield* is the broader project name; this release does not provide harvest-weight, fruit-mass or per-area yield measurements, and the count labels should therefore be interpreted as image-level berry counts rather than yield estimates. The images include 424 smartphone greenhouse images, 67 video-derived frames, and 23 DJI Fly video-frame samples, providing a reproducible dataset for ripeness segmentation, berry counting, and class-imbalance analysis in controlled-environment blueberry production.

*Corresponding author: yusra.abdulrahman@ku.ac.ae

## Background & Summary

Visual cues remain central to ripeness decisions in blueberry production [1, 2, 3, 6, 7]. A scout or grower is not only asking whether an individual berry is blue; they are assessing clusters that may contain green, purple, ripe and over-ripe fruit at the same time. The resulting visual problem is dense, multiclass and spatially mixed, which makes pixel-level masks and image-level counts more reusable than image-level labels alone [8, 9, 10].

Recent agricultural vision datasets demonstrate the value of releasing images alongside masks, counts, metadata, and fixed validation protocols [17, 18, 19, 20, 21]. Comparable open dataset descriptors span fruit ripeness grading, crop mapping and plant phenotyping tasks [22, 23, 24, 25, 26], along with leaf-, kernel- and field-scale image collections [27, 28, 29, 30, 31]. AerialYield-B$^2$D follows this dataset-descriptor model while focusing specifically on greenhouse blueberry ripeness stages and close-range/video-frame imagery rather than broad crop inventories or yield tables; it complements existing blueberry image collections [4, 5, 11] and broader fruit detection and segmentation benchmarks [12, 13, 14].

Our contribution is AerialYield-B$^2$D, a focused greenhouse blueberry dataset with five ripeness-stage masks and image-level count metadata. B$^2$D denotes *BlueBerry Dataset*; *AerialYield* refers to the broader research project, not to a claim that every image is aerial or that yield measurements are included. The dataset is explicitly presented as a real-image resource. Synthetic augmentation was explored only as a training-side imbalance experiment and is kept separate from the primary data records. The release record, counts, metadata and technical-validation results reported here refer to the 514 real RGB images (Figure 1).

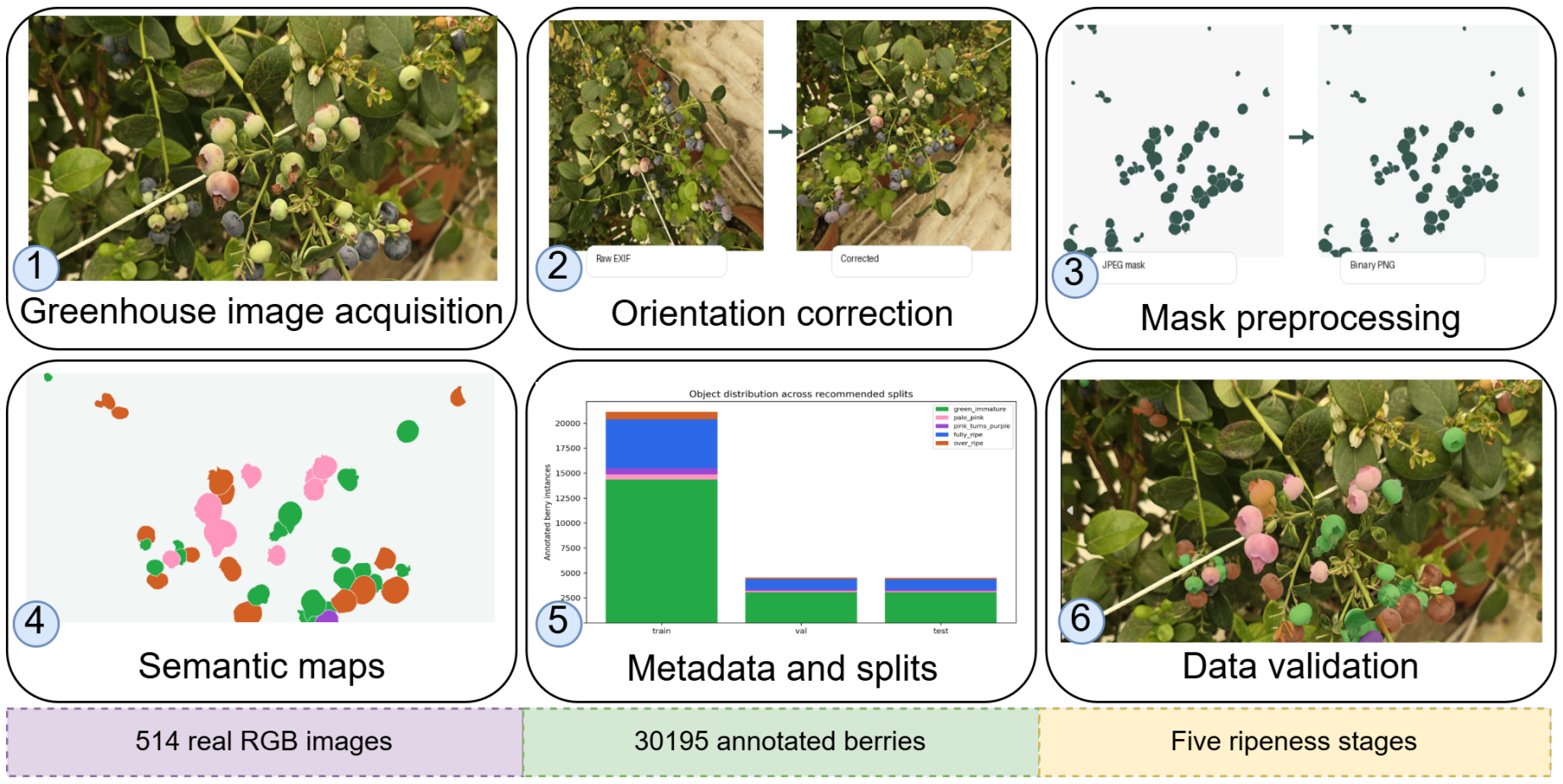

*Figure 1. Dataset curation workflow for AerialYield-B$^2$D. The numbered panels show the six curation stages: (1) greenhouse image acquisition, (2) EXIF orientation correction, (3) mask preprocessing, (4) semantic map generation,*

*(5) metadata and split construction, and (6) data validation. The bottom band summarizes the 514 real RGB images, 30,195 annotated berries and five ripeness stages.*

# Methods

## Image acquisition

The current release contains 514 RGB images of greenhouse blueberry plants (Figure 2). The original subset comprises 424 close-range smartphone images with embedded device metadata, mainly released at 3000 x 4000 pixels after Exchangeable Image File Format (EXIF) orientation correction. The added 90-image subset comprises 67 high-resolution video-derived frames at 4320 x 7680 pixels and 23 DJI Fly video-frame samples at 1280 x 720 pixels. Representative smartphone and DJI Fly video-frame examples are shown in Figure 3. The added frames did not contain embedded EXIF camera fields in the recovered JPEG files; their dates and DJI Fly provenance were therefore recorded from filenames and retained as filename-derived metadata in the release manifest. Table 9 summarizes the source modalities and metadata provenance.

The acquisition dates are from 12 February 2026 to 30 April 2026. Images were collected at Silal Al Foah Farm, Al Ain, Abu Dhabi, United Arab Emirates, from a 0.5-hectare greenhouse blueberry block (Figure 2). The block contained 2,743 blueberry plants of variety A-B, transplanted on 28 March 2023 and distributed across 13 spans. The release metadata records the image source set, source modality, embedded EXIF fields where available, filename derived date/time fields where applicable and image geometry. Including both smartphone close-range images and DJI/video-frame imagery is intentional. The two sources capture complementary acquisition conditions: smartphone images emphasize berry colour, wax bloom, fine boundaries and close cluster structure, whereas DJI/video-frame samples add wider canopy views, denser plant-scale context, stronger scale variation and frame-extraction conditions that are closer to practical greenhouse scouting. This mixture, therefore, supports testing whether ripeness segmentation and counting methods remain reliable across realistic image sources used in handheld inspection, robotic scouting and overhead monitoring. The source modality is explicitly recorded in dataset_manifest.csv, allowing users to train on the full release, evaluate modality-specific performance, or restrict analyses to a single acquisition source.

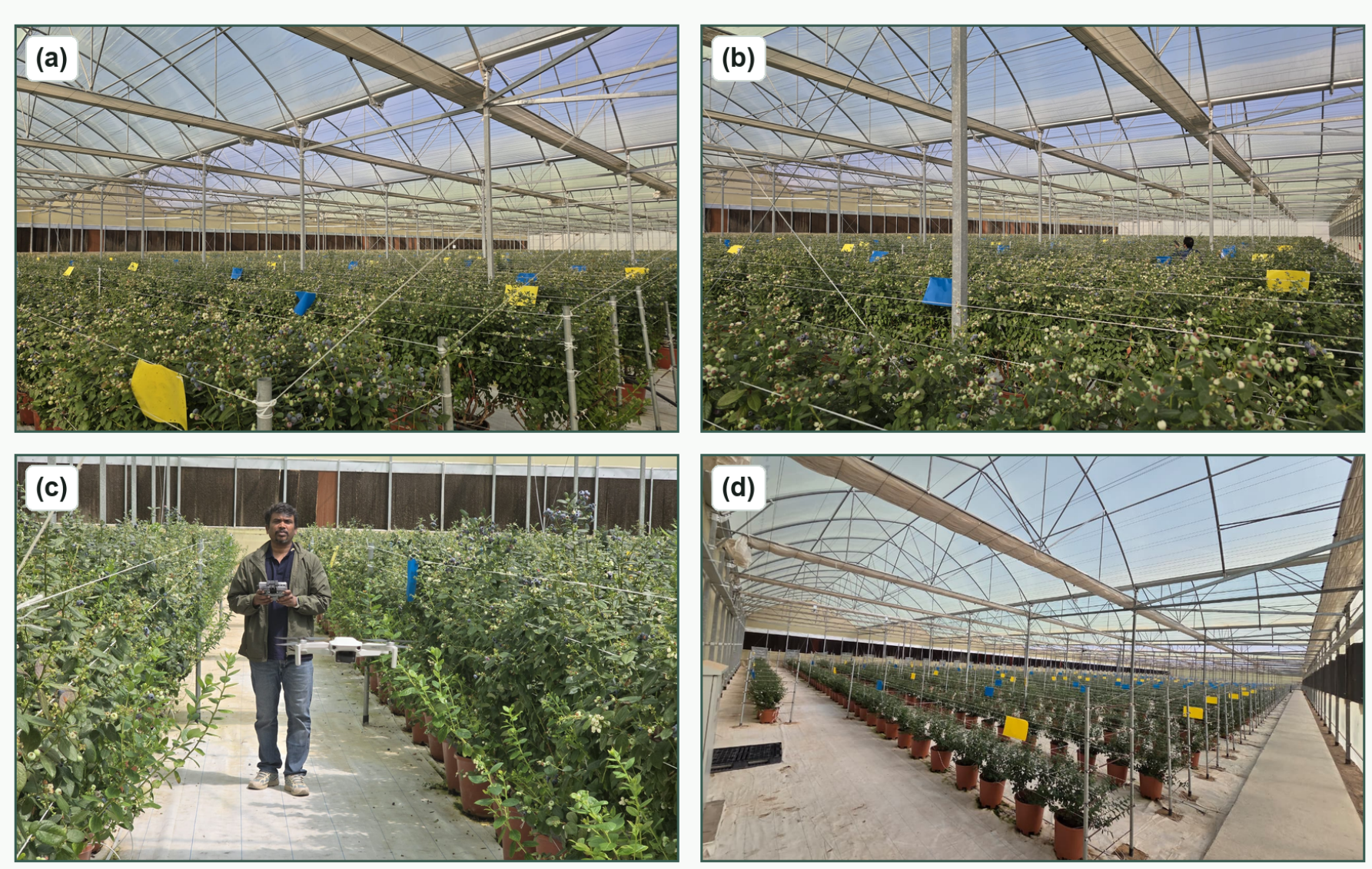

*Figure 2. Greenhouse data-collection environment at Silal Al Foah Farm, Al Ain, Abu Dhabi, United Arab Emirates. Images were collected from a 0.5-hectare greenhouse blueberry block. (a,b) show wide interior views of the trellised potted rows from different angles, (c) shows drone-assisted acquisition with an operator piloting the DJI drone between rows, and (d) shows an aisle-length view of the pot layout receding down the greenhouse. Site information: https://www.silal.ae/.*

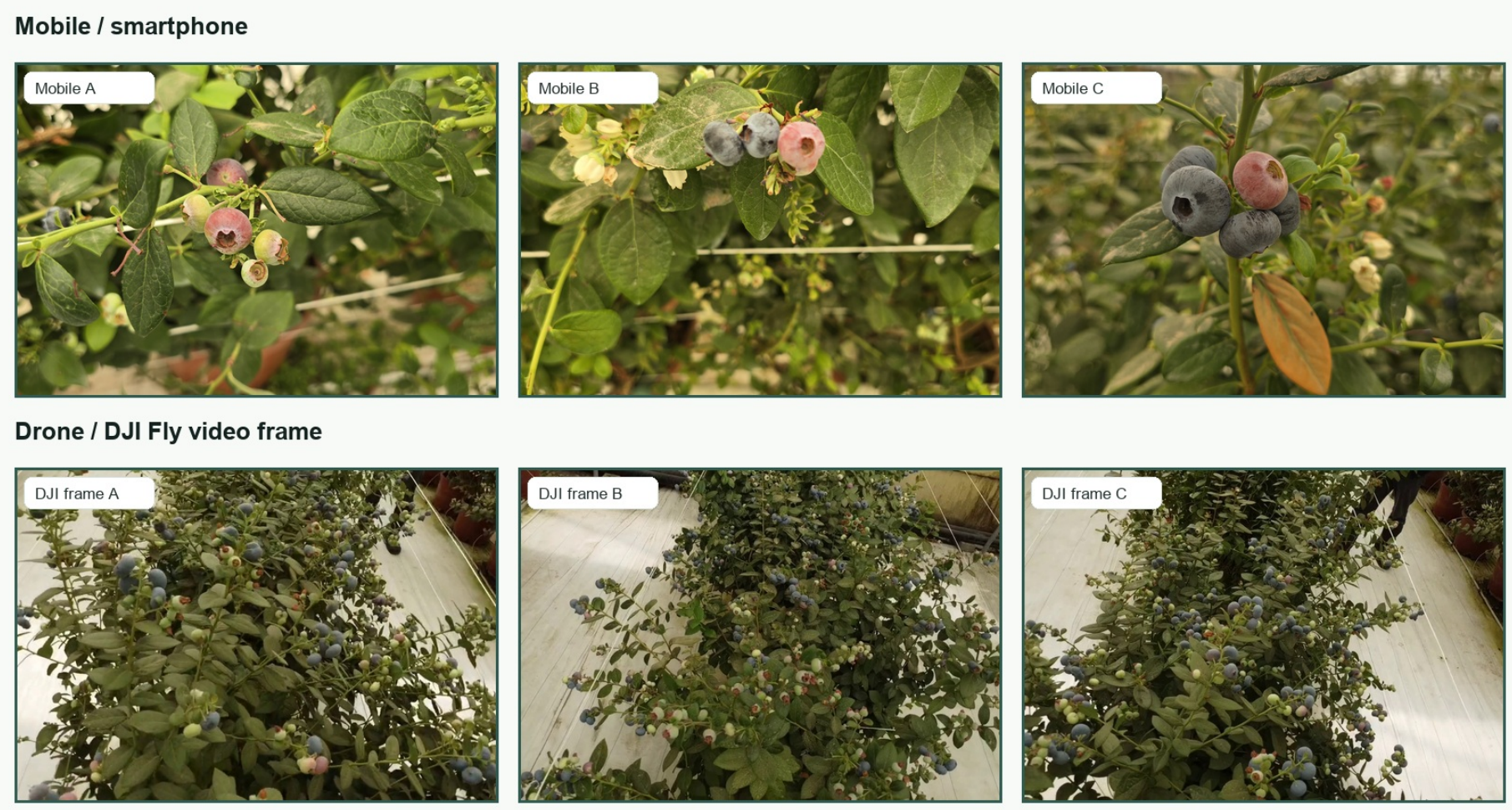

*Figure 3. Example images from AerialYield-B$^2$D. The top row shows smartphone examples, and the bottom row shows DJI Fly video-frame examples. Sample letters are used only for visual comparison; filenames and full metadata are provided in dataset_manifest.csv.*

### Data collection workflow

To make the acquisition procedure explicit, data collection was treated as a scene-level sampling task rather than a class-balanced object-capture task. Images were collected from greenhouse blueberry plants while keeping fruit clusters in their natural canopy context, including occlusion, foliage, lighting variation and mixed ripeness stages. The resulting class imbalance, therefore, reflects the crop state captured during acquisition rather than an artificial balancing target.

- Step 1 - Identify greenhouse blueberry scenes containing visible berry clusters, including scenes with mixed ripeness stages and natural occlusion.
- Step 2 - Acquire close-range RGB images and video-frame samples, retaining source JPEG files, embedded EXIF metadata when present and filename-derived acquisition metadata when EXIF fields are absent.
- Step 3 - Preserve greenhouse context rather than isolating individual berries, so that downstream users can evaluate segmentation under realistic cluster, foliage and lighting conditions.
- Step 4 - Assign images to manual Computer Vision Annotation Tool (CVAT) [32] annotation batches, trace individual berry boundaries with polygons and assign the ripeness attribute for each annotated berry.
- Step 5 - Export class-specific and overall masks, then apply orientation correction, thresholded PNG mask conversion and semantic-map generation.
- Step 6 - Generate image-level counts, source metadata, SHA-256 hashes, fixed splits, quality reports and baseline-validation artifacts for release.

This workflow complements the visual curation summary in Figure 1 and explains why the released data are suited to segmentation, count-label benchmarking and ripeness-distribution analysis while retaining natural greenhouse clutter, occlusion and mixed-stage clusters.

## Ripeness taxonomy

The labels follow a practical colour-based ripeness taxonomy rather than a laboratory maturity evaluation. Class names, label identifiers and display colours are defined in class_map.json and used consistently across semantic maps, binary masks, overlay figures (Figure 4) and Table 1.

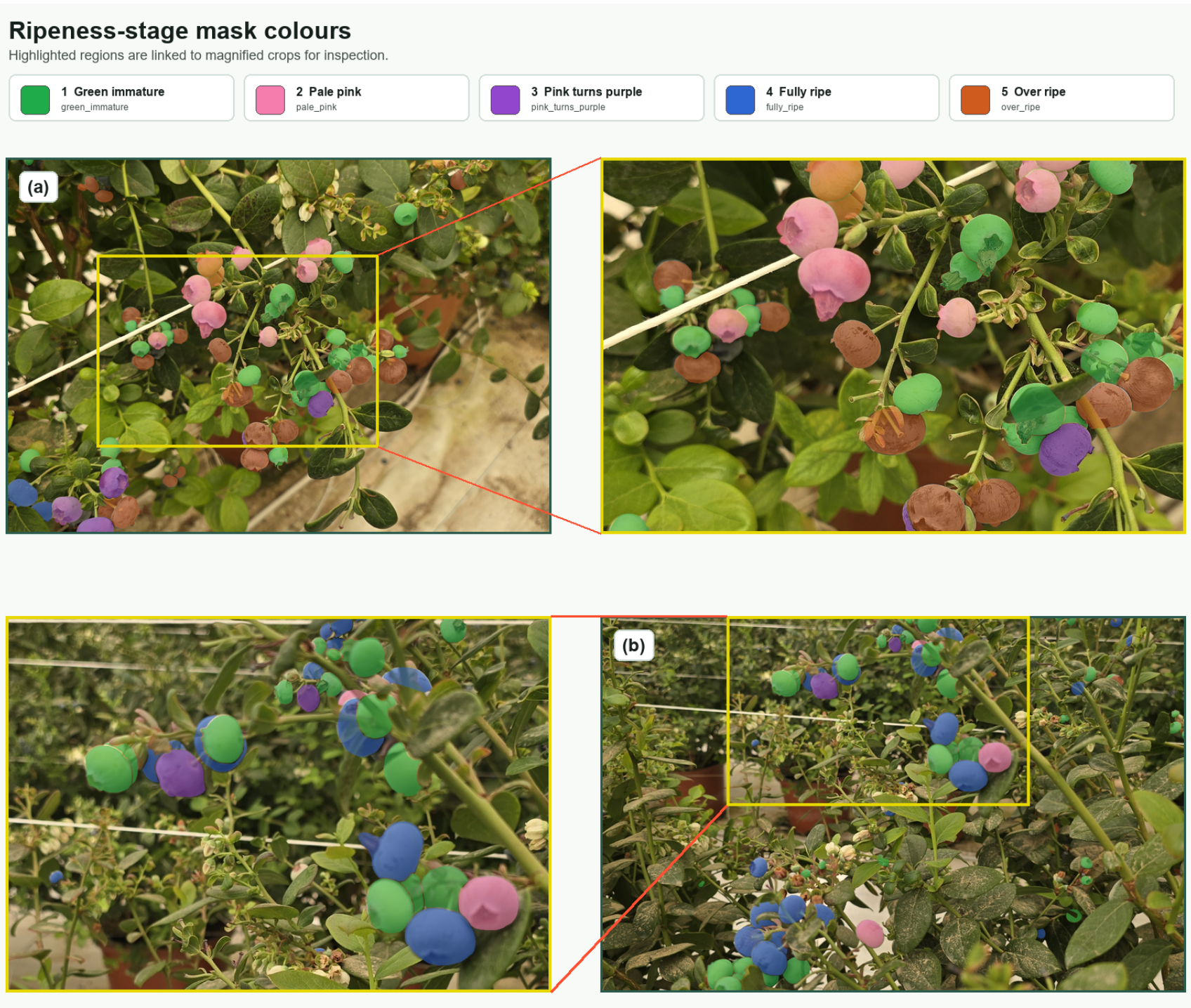

*Figure 4. Ripeness-stage mask colours and zoomed overlay inspection. The enlarged panels show selected crop regions linked by diagonal connector lines, making the five mask colours and local annotation boundaries easier to inspect.*

*Table 1. Semantic label map, class definitions and mask colours used in the release package.*

| label id | class name | display name | semantic value | notes |
|---|---|---|---|---|
| 0 | background | Background | 0 | No overlay colour; non-berry pixels. |
| 1 | green_immature | Green immature | 1 | Mask colour: green (#23AA46; RGB 35, 170, 70). |

| label id | class name | display name | semantic value | notes |
|---|---|---|---|---|
| 2 | pale_pink | Pale pink | 2 | Mask colour: pale pink (#FF96BE; RGB 255, 150, 190). |
| 3 | pink_turns_purple | Pink turns purple | 3 | Mask colour: purple (#9B46D2; RGB 155, 70, 210). |
| 4 | fully_ripe | Fully ripe | 4 | Mask colour: blue (#2D69E6; RGB 45, 105, 230). |
| 5 | over_ripe | Over ripe | 5 | Mask colour: orange (#D25F23; RGB 210, 95, 35). |
| 255 | ignore | Ignore | 255 | Ignore value for class-overlap pixels; not a ripeness colour. |

## Annotation products

The annotations were produced manually in CVAT by trained annotators working in batches. The work followed a plain, auditable sequence: sample annotation, feedback, production annotation, quality checks, output checks, delivery, rework where needed and final approval. This step-by-step process was used to check early labelling decisions before the full dataset was completed.

The annotation task was instance-aware. Annotators identified individual blueberries, traced fruit boundaries with polygons and assigned the appropriate ripeness attribute. No automatic pre-labelling or model-generated annotations were used. After production, the quality-control team reviewed the annotated data before export.

## Preprocessing pipeline

1. Scan raw RGB image and mask folders.
2. Apply EXIF orientation correction to RGB images.
3. Convert source JPEG or PNG masks to thresholded PNG binary masks using threshold >127.
4. Preserve per-class binary masks and overall masks.
5. Build semantic label maps with label IDs 0-5 and ignore value 255 for class-overlap pixels.
6. Compute per-image object counts from annotation exports and connected-component statistics from masks.
7. Compute SHA-256 hashes for traceability.
8. Generate deterministic train, validation and test splits.
9. Write metadata tables, data dictionaries, validation figures and audit reports.

The main release-building script is scripts/prepare_scientific_data_release.py. The 514-image staging script, scripts/prepare_combined_514_data_root.py, combines the original 424-image subset with the added 90-image frame subset before release generation. Benchmark annotations are prepared with scripts/fresh_prepare_annotations.py, and model validation is run with scripts/fresh_run_task.py or scripts/fresh_run_all.py.

## Recommended data split

We created a deterministic 360/77/77 train/validation/test split. The split balances image counts and approximates object-count balance across ripeness stages. It is included mainly to make benchmark comparisons reproducible; users may define other splits when their study design requires it, but should avoid image leakage across train, validation and test subsets.

## Synthetic augmentation experiments

Synthetic augmentation was tested as a training-only response to class imbalance. It is not part of the primary real dataset record and should not be counted as additional observations in the Data Records. The real-image validation tables below therefore report only the 514-image release split.

## Ethics statement

The dataset contains plant and crop images. It does not contain human participants, human-derived data or animal subjects.

# Data Records

The release package contains 4,142 files and occupies approximately 1.88 GiB. The expected repository or institutional-drive record should contain the release folder described in Table 2, including the dataset files, metadata, split files, technical-validation reports, figures and checksum files.

***Table 2. Expected repository folder structure and file contents.***

| folder | contents |
|---|---|
| dataset/images | EXIF-corrected RGB JPEG images. |
| dataset/masks_binary/<class_name> | Lossless PNG binary masks for each of the five ripeness classes. |
| dataset/masks_semantic | Single-channel semantic label maps using values 0-5 and 255 for ignore pixels. |
| dataset/masks_overall | Overall berry masks independent of ripeness class. |
| metadata | Dataset manifest, image-level counts, class distribution, split summary, class map and audit tables. |
| splits | Train, validation and test split files. |
| reports | Textual validation and preprocessing reports. |
| figures | Processing, overlay, distribution, image-quality and baseline-validation figures. |

## Metadata fields

The key metadata files are plain CSV files: dataset_manifest.csv is the main entry point and records image identifiers, filenames, source set, source modality, split assignments, release-relative image and mask paths, image geometry, EXIF or filename-derived date/time fields, SHA-256 hashes, quality metrics, object counts and mask statistics. class_distribution.csv, split_summary.csv and image_level_counts.csv provide compact summaries for reuse and paper reporting.

# Data Overview

The following summaries provide readers with a quick overview of the annotation volume and the recommended benchmark split. Table 3 summarizes object counts and image coverage, Table 4 reports mask-pixel area and connected components, and Table 5 reports the fixed train, validation and test split. Figure 5 and Figure 6 provide the corresponding object-distribution views.

*Table 3. Class object counts and image coverage.*

| class id | class name | objects | object percent | images with objects |
|---|---|---|---|---|
| 1 | green_immature | 20,525 | 67.97 | 511 |
| 2 | pale_pink | 736 | 2.44 | 339 |
| 3 | pink_turns_purple | 883 | 2.92 | 318 |
| 4 | fully_ripe | 7,002 | 23.19 | 493 |
| 5 | over_ripe | 1,049 | 3.47 | 128 |

*Table 4. Mask-pixel and connected-component distribution.*

| class id | class name | mask pixels | mask area percent | connected components |
|---|---|---|---|---|
| 1 | green_immature | 201,246,760 | 2.74 | 16,389 |
| 2 | pale_pink | 15,104,904 | 0.21 | 731 |
| 3 | pink_turns_purple | 17,211,448 | 0.23 | 898 |
| 4 | fully_ripe | 97,930,816 | 1.34 | 6,515 |
| 5 | over_ripe | 15,997,775 | 0.22 | 879 |

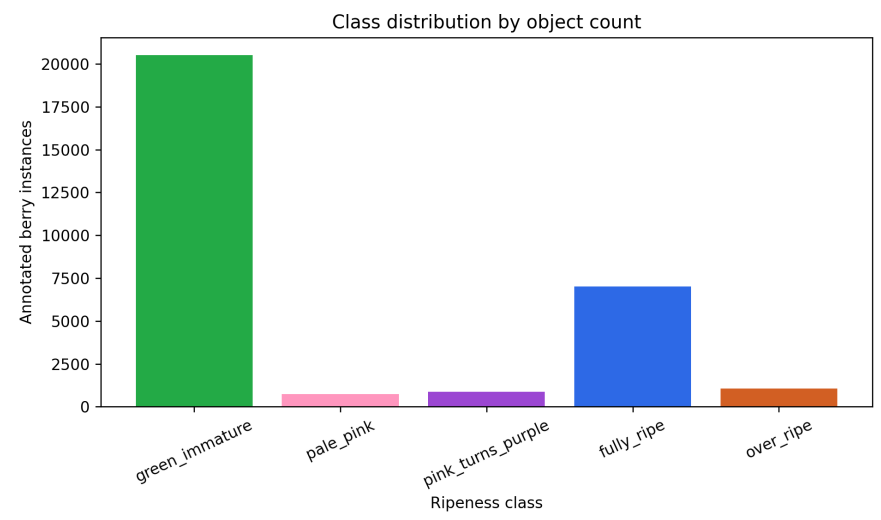

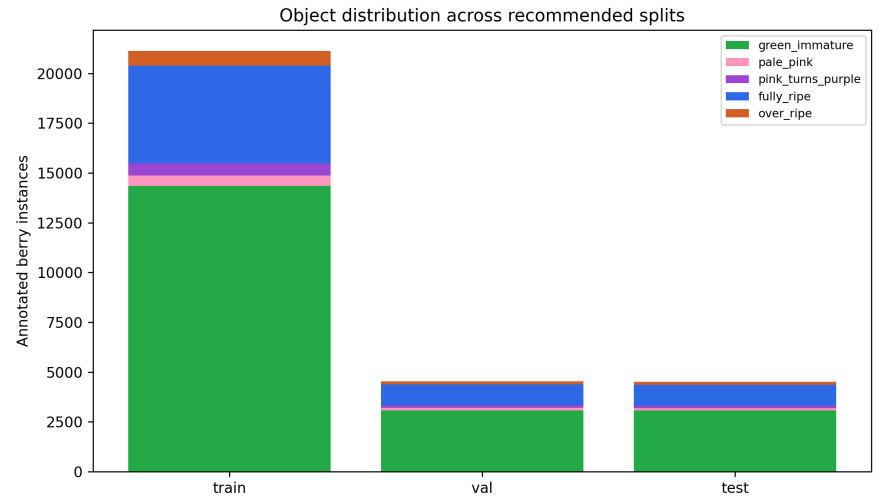

*Figure 5. Object-count distribution across the five ripeness classes in the 514-image release. The distribution documents persistent class imbalance, especially for pale-pink, pink-turns-purple and over-ripe berries.*

*Figure 6. Object distribution across the recommended train, validation and test splits. The fixed 360/77/77 split supports comparable benchmarking.*

*Table 5. Recommended train, validation and test split summary.*

| split | images | total objects | green immature objects | pale pink objects | pink turns purple objects | fully ripe objects | over-ripe objects |
|---|---|---|---|---|---|---|---|
| train | 360 | 21,127 | 14,358 | 514 | 620 | 4,898 | 737 |
| val | 77 | 4,543 | 3,086 | 111 | 131 | 1,060 | 155 |
| test | 77 | 4,525 | 3,081 | 111 | 132 | 1,044 | 157 |

# Technical Validation

## Completeness and file integrity

The preprocessing audit checked 514 RGB image-mask groups. Every RGB image has one overall mask and five class-specific masks. The release image SHA-256 hashes are unique for all 514 images, filenames are unique, and no missing image-mask pairs were found during preprocessing.

## Image geometry and orientation

Images were EXIF-transposed prior to release so that RGB pixels directly correspond to masks. The release contains 423 images at 3000 x 4000 pixels, 1 image at 4000 x 3000 pixels, 67 video-derived frames at 4320 x 7680 pixels and 23 DJI Fly video-frame samples at 1280 x 720 pixels. The audit recorded EXIF orientation value 6 for 423 smartphone images and orientation value 1 for the remaining 91 images.

## Mask conversion and overlap audit

The source masks include JPEG-derived masks from the original 424-image subset and binary PNG masks from the added 90-image annotation subset. The release converts all masks to thresholded binary PNG using threshold >127. The audit logged 32,768,752 ambiguous edge or compression pixels before thresholding, 74 images with class-overlap pixels after thresholding and 237 images with any class-union versus overall-mask mismatch. Overlap pixels are encoded as value 255 in semantic maps and should be treated as ignored pixels during semantic-segmentation training.

## Qualitative overlay inspection

Qualitative validation was conducted by visual overlay inspection. Representative greenhouse RGB images, along with their ripeness masks, are shown in Figure 4. We checked the RGB-mask alignment after EXIF orientation correction, the visual plausibility of ripeness labels, the preservation of plant geometry, and challenging cases such as occlusion, dense clusters, small berries, shadows and overlapping annotations. These overlays are not statistical evidence but aids to inspection and are important for detecting obvious annotation or orientation failures.

## Image-quality screening

Brightness, contrast, saturation and Laplacian sharpness were measured on downsampled, EXIF-oriented images to screen for gross acquisition problems. Across the 514 images, mean brightness averaged 109.72, brightness standard deviation averaged 52.97, Laplacian sharpness variance averaged 1286.73 and saturation averaged 125.79; the distributions are shown in Figure 7.

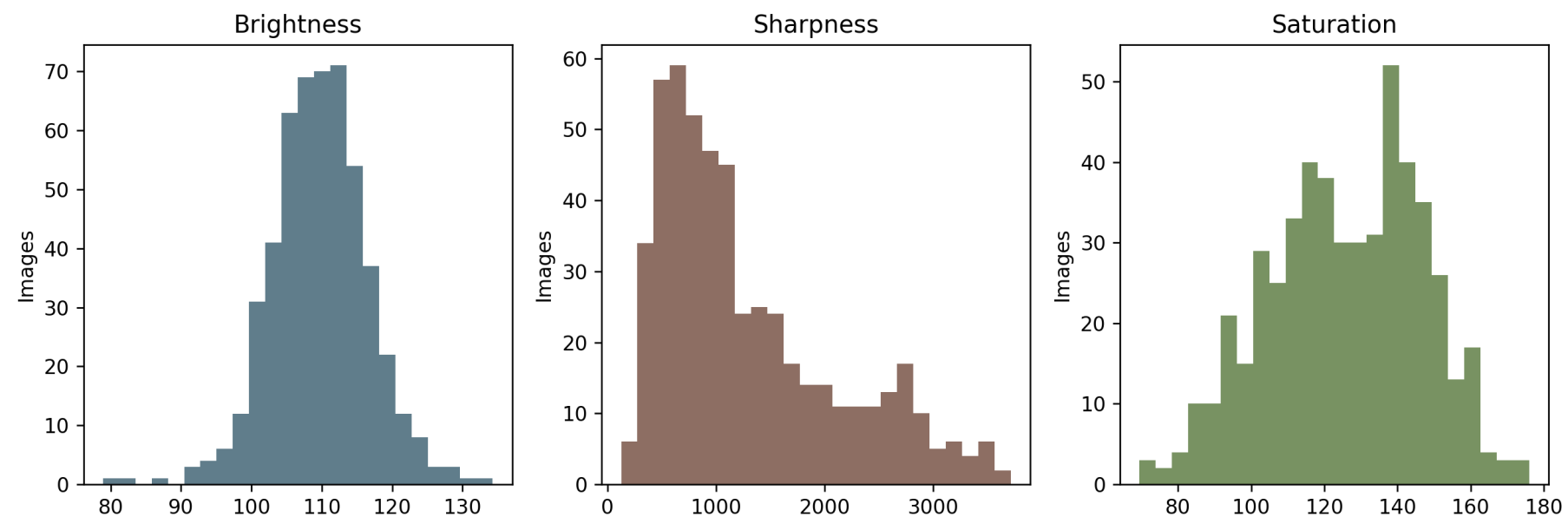

*Figure 7. Image-quality screening distributions for the 514 released RGB images. Panels show brightness mean, Laplacian sharpness variance, and saturation mean*

*measured on downsampled EXIF-oriented images; the distributions support the absence of grossly blank or unusable image groups.*

## Counting validation

Because the release includes image-level berry counts, we added a held-out counting baseline rather than leaving counting as an unsupported use claim. An EfficientNetV2-S regression model trained on the recommended split achieved MAE = 13.86 berries, RMSE = 22.23 berries and $R^2 = 0.858$ on the held-out test images (Table 8 and Figure 8b). These results demonstrate that the count labels can support reproducible benchmarking while also showing that dense greenhouse scenes remain challenging. These count predictions are not converted into yield because the release does not include fruit mass, harvested weight per plant, harvest totals or calibrated plot-area yield measurements.

## Baseline semantic-segmentation validation

Semantic-segmentation validation was repeated on the 514-image split using an FPN-style segmentation model [15] with a ConvNeXtV2-T backbone [16]. We treat these runs as technical validation rather than as a claim that the chosen architecture is optimal. The baseline achieved foreground mIoU = 0.4681, foreground Dice = 0.6128 and pixel accuracy = 0.9836 on the held-out test split (Table 6, Table 7 and Figure 8a).

*Table 6. Baseline semantic-segmentation validation on the 514-image release split.*

| method | train images | val images | test images | best epoch | best val metric | mIoU foreground | Dice foreground | pixel accuracy |
|---|---|---|---|---|---|---|---|---|
| FPN ConvNeXtV2-T | 360 | 77 | 77 | 12 | 0.4815 | 0.4681 | 0.6128 | 0.9836 |

*Table 7. Per-class held-out semantic-segmentation metrics for the FPN ConvNeXtV2-T baseline.*

| class id | class name | IoU | Dice | split |
|---|---|---|---|---|
| 0 | background | 0.9871 | 0.9935 | held-out test |
| 1 | green_immature | 0.6226 | 0.7674 | held-out test |
| 2 | pale_pink | 0.4619 | 0.6320 | held-out test |

| class id | class name | IoU | Dice | split |
|---|---|---|---|---|
| 3 | pink_turns_ purple | 0.4851 | 0.6533 | held-out test |
| 4 | fully_ripe | 0.6404 | 0.7808 | held-out test |
| 5 | over_ripe | 0.1303 | 0.2305 | held-out test |

*Table 8. Held-out image-level berry-count regression baseline on the 514-image release split.*

| method | train images | val images | test images | best epoch | val MAE | test MAE | test RMSE | test $R^2$ |
|---|---|---|---|---|---|---|---|---|
| EfficientNetV2-S regression | 360 | 77 | 77 | 7 | 13.37 | 13.86 | 22.23 | 0.858 |

*Table 9. Acquisition source and metadata summary for the 514-image release.*

| source modality | images | metadata source | image geometry | notes |
|---|---|---|---|---|
| smartphone | 424 | embedded EXIF | 3000 x 4000 after EXIF correction for 423 images; one 4000 x 3000 image | original close-range subset |
| video_frame | 67 | filename-derived | 4320 x 7680 | added 2026-02-12 video-frame subset |
| drone_video_frame | 23 | filename-derived DJI Fly pattern | 1280 x 720 | added 2026-02-17 DJI Fly video-frame subset |

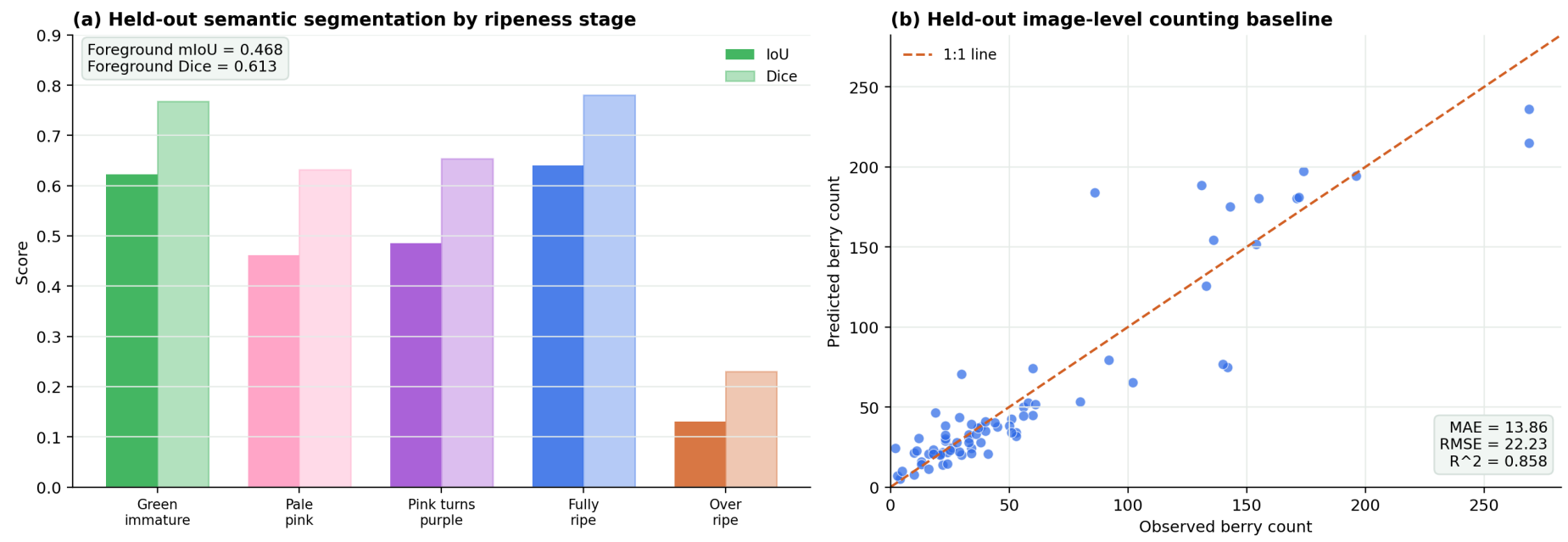

*Figure 8. Model-based technical validation on the 514-image release split. Panel (a) reports held-out per-class semantic-segmentation IoU and Dice. Panel (b) compares observed image-level berry counts with EfficientNetV2-S count predictions on held-out test images.*

# Usage Notes

For semantic segmentation, use dataset/images as inputs and dataset/masks_semantic as targets. Treat value 255 as an ignore label. For multilabel segmentation, use the class-specific binary masks in dataset/masks_binary. For counting, use image_level_counts.csv or the count fields in dataset_manifest.csv. For detection or instance-level crop experiments, use the generated COCO/YOLO exports with the caveat that connected components can merge in very dense clusters.

The size of the dataset should be assessed not only by the number of images it contains but also by the volume of dense annotations. The 514 real RGB images contain 30,195 annotated berry instances, including 736 pale-pink, 883 pink-turns-purple and 1,049 over-ripe instances. Rare classes remain limited compared with green immature and fully ripe fruit, so class imbalance should be considered part of the benchmark rather than hidden.

The recommended split is provided for reproducible benchmarking. Users creating new splits should maintain image-level separation and document their split-generation process. AerialYield-B$^2$D should be used as a close-range/video-frame greenhouse blueberry ripeness resource; it does not provide field-scale yield measurements.

# Data Availability

The dataset described in this Data Descriptor will be deposited in [Zenodo / Figshare]. The repository record should contain the release folder described in the Data Records section, including RGB images, class-specific masks, semantic masks, metadata files, split files, validation reports, figures and SHA-256 checksum files.

## Code Availability

Custom preprocessing, release-generation, synthetic-augmentation and validation scripts are available at https://github.com/iyyakuttiiyappan/Blueberry-Ripeness-Dataset. The repository includes the scripts, Python dependency file, documentation, configuration files and package modules required by the validation workflow. The repository also specifies the versions or archived releases used to generate the dataset and validation outputs.

## Author Contributions

I.I.G., and A.A.: Conceptualization, Methodology, Software, Data Curation, Investigation, Writing.

M.O., I.H., and Y.A.: Supervision, Writing - Reviewing and Editing, Project Administration.

## Competing Interests

The authors declare no conflict of interest.

## Acknowledgements

This research was funded by Khalifa University of Science and Technology through the Silal: *Smart AgriTech Drone for Automated Yield Estimation and Fruit Ripeness Analysis Using Advanced AI-Powered Precision Agriculture Techniques*, under Project ID: KU-EXT-2025-8475000025.